\documentclass[review]{elsarticle}

\usepackage[OT4]{fontenc}

\usepackage[utf8]{inputenc} 
\usepackage{lineno,hyperref}
\usepackage{adjustbox}
\usepackage{algorithm}
\usepackage{graphicx}
\usepackage[noend]{algpseudocode}
\usepackage{amsthm,amsmath}
\usepackage{amssymb}
\usepackage{array}
\usepackage{color}
\usepackage{enumerate}
\usepackage{multirow}
\usepackage{caption}
\usepackage[section]{placeins}
\modulolinenumbers[5]

\journal{ }

\newcommand{\cmt}[1]{{\color{red} #1}}

\newcommand{\Rexp}{R_\oplus}
\newcommand{\Ap}{A_\oplus}
\newcommand{\Am}{A_\ominus}
\newcommand{\Cp}{C_\oplus}
\newcommand{\Cm}{C_\ominus}
\newcommand{\mincov}{\textit{mincov}}

\newcommand{\Epref}{\Sigma_\textrm{pref}}
\newcommand{\Eauto}{\Sigma_\textrm{auto}}
\newcommand{\Ypref}{\Upsilon_\textrm{pref}}
\newcommand{\Yauto}{\Upsilon_\textrm{auto}}

\newcommand{\ii}[1]{\textit{#1}}

\newcommand{\IF}{\textbf{IF }}
\newcommand{\THEN}{\textbf{ THEN }}
\newcommand{\AND}{\ \pmb{\wedge}\ }

\newcommand{\e}{\phantom{0}}

\newcolumntype{R}[2]{%
	>{\adjustbox{angle=#1,lap=\width-(#2)}\bgroup}%
	l%
	<{\egroup}%
}
\newcommand*\rot{\multicolumn{1}{R{45}{1em}}}

\begin{document}

\begin{frontmatter}

\title{GuideR: a guided separate-and-conquer rule learning in classification, regression, and survival settings}


\author[polsl,emag]{Marek Sikora\corref{mycorrespondingauthor}}
\ead{marek.sikora@polsl.pl}

\author[polsl,emag]{\L{}ukasz Wr\'obel\corref{mycorrespondingauthor}}
\ead{lukasz.wrobel@polsl.pl}

\author[polsl]{Adam Gudy\'s\corref{mycorrespondingauthor}}
\ead{adam.gudys@polsl.pl}

\cortext[mycorrespondingauthor]{Corresponding author}
\address[polsl]{Institute of Informatics, Silesian University of Technology, Akademicka 16, 44-100 Gliwice, Poland}
\address[emag]{Institute of Innovative Technologies, EMAG, Leopolda 31, 40-189 Katowice, Poland}

\begin{abstract}
This article presents GuideR, a user-guided rule induction algorithm, which overcomes the largest limitation of the existing methods—--the lack of the possibility to introduce user's preferences or domain knowledge to the rule learning process. Automatic selection of attributes and attribute ranges often leads to the situation in which resulting rules do not contain interesting information. We propose an induction algorithm which takes into account user’s requirements. Our method uses the sequential covering approach and is suitable for classification, regression, and survival analysis problems. The effectiveness of the algorithm in all these tasks has been verified experimentally, confirming guided rule induction to be a powerful data analysis tool.
\end{abstract}

\begin{keyword}
Rule induction \sep User-guided rule induction \sep Semi-automatic rule induction \sep Classiffication \sep Regression \sep Survival analysis
\end{keyword}

\end{frontmatter}


\section{Introduction}

Sequential covering rule induction algorithms can be used for both, predictive and descriptive purposes \cite{blaszczynski2011,furnkranz1999,grzymala2003,kaufman1991}. In spite of the development of increasingly sophisticated versions of those algorithms \cite{liu2018induction, valmarska2017}, the main principle remains unchanged and involves two phases: rule growing and rule pruning. In the former, the elementary conditions are determined and added the rule premise. In the latter, some of these conditions are removed.

In comparison to other machine learning methods, rule sets obtained by sequential covering algorithm, also known as separate-and-conquer strategy (SnC), are characterized by good predictive as well as descriptive capabilities. Taking into consideration only the former, superior results can often be obtained using other methods, e.g. neural-fuzzy networks, support vector machines, or ensemble of classifiers \cite{boser1992,czogala2000,rokach2010,siminski}, especially ensemble of rules \cite{dembczynski2010}. However, data models obtained this way are much less comprehensible than rule sets.

In the case of rule learning for descriptive purposes, the algorithms of association rule induction \cite{agrawal1994,kavsek2006,stefanowski2001} or subgroup discovery \cite{lavravc2004, valmarska2017}, are applied. The former leads to a very large number of rules which must then be limited by filtering according to rule interestingness measures \cite{geng2006,greco2016, bayardo}. Nevertheless, rule sets obtained by subgroup discovery are characterized by worse predictive abilities than those generated by the standard sequential covering approach.

Therefore, if creating a prediction system with comprehensible data model is the main objective, the application of sequential covering rule induction algorithms provides the most sensible solution.

In the works \cite{wrobel2017,wrobel2016,sikora2012, sikora2013data}, we have presented and confirmed on dozens of benchmark datasets the effectiveness of our version of the sequential algorithm for generating classification, regression, and survival rules.
This article presents the semi-interactive version of that algorithm, which overcomes the largest limitation of the existing rule induction methods---the lack of the possibility to introduce user's knowledge (or expert's knowledge) to the learning process. Automatic selection of attributes and attribute ranges often leads to the situation in which induced rules do not contain the most important information from the user's point of view.
We propose a rule induction algorithm which takes into account user’s requirements. The possibility to specify the initial set of rules, preferred and forbidden conditions/attributes, etc., together with the multiplicity of options and modes, makes our algorithm the most flexible solution for user-guided rule induction. It allows testing various hypotheses concerning data dependencies which are expected or of interest. In particular, the algorithm enables making such hypotheses more specific or more general.

The effectiveness of the guided (semi-automatic) rule induction has been investigated on three test cases concerning various data analysis tasks. Classification was illustrated by the problem of predicting seismic hazards in coal mines (\textit{seismic-bumps} dataset \cite{Sikora2010}); regression---the problem of methane forecasting (\textit{methane} dataset~\cite{githubMethane}); survival analysis---the problem of analysing factors which impact patients’ survival following bone marrow transplants (\textit{BMT-Ch} dataset \cite{kalwak2010,sikora2013}).
	

The paper is organized as follows: Section~\ref{sec:related} concerns overview of works in the area of user-guided rule induction. Section~\ref{sec:methods} presents the algorithm for induction of classification, regression, and survival rules, with a special stress put on the semi-automatic capabilities. Section~\ref{sec:results} is devoted to the analysis of the test cases, together with a discussion of obtained results. Section~\ref{sec:conclusions} contains a summary and conclusions.

GuideR software as well as the datasets used in this article are available at \url{https://github.com/adaa-polsl/GuideR} or \url{http://www.adaa.polsl.pl}. All the datasets had been proposed by the authors of this article. The \textit{sesimic-bumps} dataset is also available in the UCI repository.

\section{Related work}
\label{sec:related}

The induction of classification rules with sequential covering approach has been known for many years \cite{furnkranz1999,grzymala2003,kaufman1991,clark1989}. As it proved its effectiveness in terms of both, the classification accuracy as well as the descriptive abilities of the induced rules (e.g. \cite{sikora2011pattern, moshkov2008, tsumoto2004}), a number of interesting extensions of this approach have been presented \cite{napierala2012,huhn2009,mozina2007,riza2014,sikora2013redef,liu2018induction,valmarska2017}. In contrast, rule induction algorithms have been rarely applied to the regression and survival analysis, although the comprehensibility of resulting data models in these problems is often a key issue.

Regression rules can be straightforwardly derived from regression trees such as CART \cite{breiman1984} and M5 \cite{quinlan1992} by generating one rule for each path from the root of the tree to its leaf. These algorithms use the divide-and-conquer strategy. The other approach to the regression rule induction is to use a generalization of sequential covering (e.g., PCR\cite{vzenko2005}, rule list \cite{janssen2011}). The work of Janssen and F\"{u}rnkranz \cite{janssen2011} describing the dynamic reduction of the regression problem to the classification is of particular importance in the context of the results presented in this paper. The most advanced methods of regression rule induction are based on ensemble techniques (e.g., RuleFit \cite{friedman2008}, RegENDER \cite{dembczynski2008}). To supervise the induction of subsequent rules, these algorithms apply gradient-based optimization methods. The resulting rule sets are characterized by good prediction quality, though they are usually composed of a large number of rules.

Equally few attempts have been made to apply rules to survival analysis.
Pattaraintakorn and Cercone \cite{pattaraintakorn2008} described the rough set-based intelligent system for analyzing survival data. Another approach employing rough sets was presented by Bazan et al.\cite{bazan2002}. The idea was to divide examples into three decision classes on the basis of a prognostic index (PI) calculated with a use of the Cox's proportional hazard model. The division of survival dataset into three classes was also made by Sikora et al.\cite{sikora2013survival}, who applied the rule induction algorithm to the analysis of patients who underwent a bone marrow transplantation. The dataset was divided in the following groups: patients who underwent transplantation at least five years before, patients who died within five years after transplantation, and patients who are still alive but whose survival time is less than five years. Two former classes were used for rule generation, while the latter for model post-processing.
Kronek and Reddy \cite{kronek2008} proposed the extension of the Logical Analysis of Data (LAD) \cite{crama1988} for survival analysis. The LAD algorithm is a combinatorial approach to rule induction. It was originally developed for the analysis of data containing binary attributes; therefore, the discretization and binarization step is usually required. Liu et al. \cite{liu2004} adapted the patient rule induction method to the analysis of survival data. The method uses bump hunting heuristic which creates rules by searching regions in an attribute space with a high average value of the target variable. To deal with censoring, the authors use deviance residuals as the outcome variable. The idea of residual-based approach to censored outcome is derived from survival trees \cite{leblanc1992}.

In comparison to rule-based techniques, tree-based methods received much more attention in survival analysis. The key idea behind the application of tree-based techniques to survival data lies in the splitting criterion. The most popular approaches are based on residuals  \cite{leblanc1992,therneau1990} or use log-rank statistics \cite{leblanc1993} to maximize the difference between survival distributions of child nodes. We employed the latter idea in our latest separate-and-conquer rule induction algorithm which uses long-rank statistics as a rule search heuristic (rule quality measure)~\cite{wrobel2017}. We showed, that in spite of some similarities between rule and trees, our approach renders different models than the divide-and-conquer strategy of tree building.

To date, few studies have concerned rule induction algorithms which take into account user's preferences. Stefanowski and Vanderpooten \cite{stefanowski2001} presented the Explore algorithm. Based on the idea of the Apriori method, it allows the user to specify the requirements for attributes and/or their values, appearing in the rule premises.
Other studies on the induction of association rules describe examples of interactive construction of rules~\cite{rafea2004, kliegr} and the generation of the unexpected rules~\cite{padmanabhan1998}. The latter are created on the basis of user-defined templates, indicating the attributes included in the so-called typical rules. Gamberger and Lavrac \cite{gamberger2002} showed a similar proposal for the decision rule induction algorithm intended for descriptive purposes.

Adomavicius and Thazulin \cite{adomavicius2001} presented expert-driven methods of validating rule-based data models obtained via the association rule induction algorithm. The approach limits number of rules by applying rule grouping and filtering techniques which are based on the interaction with the user instead of the traditional calculation of the rules attractiveness. Balanchard et al. \cite{blanchard} proposed an interactive methodology for the visual post-processing
of association rules. It allows the user to explore large sets of rules freely by focusing his/her attention on limited subsets of rules. Both the aforementioned methods do not interfere with the induction process.

Algorithms using the paradigm of the argument-based learning \cite{mozina2007} allow the user to provide explanation for each example why it has been assigned with a particular decision class. Examples of medical applications show that this approach can significantly reduce the set of generated rules. However, the argument-based learning approach does not verify the hypotheses that represent the dependencies which, in the user’s opinion, might occur in the data. Partially, this possibility was introduced by Chen and Liu \cite{ chen2001}, where the user defines a set of rules expected to be found in the dataset. Then, the rule-based version of the C4.5 algorithm is executed and three types of rules are generated: consistent, inconsistent, and not related to the user's rules. The rule $r$ is considered to be consistent with the knowledge if in the set of defined rules, there is at least one rule $e$ such that $r$ and $e$ indicate the same decision class and a set of examples covered by $r$ is a subset of examples covered by $e$.

The IBM SPPS Modeler analytical package \cite{ibmspss} contains a module of interactive decision trees in which the user can determine the attribute and split value to be included in a given tree node. Moreover, the algorithm allows maintaining the induction of a given tree node at a specific level or starting it from a certain level when the above nodes have been defined by the user.

Even though trees can be straightforwardly translated into rules, the induction of the latter directly from data have an important advantage---the rules can be treated independently. The user or domain expert can alter existing rules or add new ones without affecting the rest of the model. The tree, in contrast, must be treated as a whole---a change of a condition in a node involves the need to modify conditions in all its child nodes. Another feature is that the divide-and-conquer tree generation strategy forbids examples to be covered by multiple rules, while the separate-and-conquer approach for rule induction lacks this limitation. This often leads to discovering stronger or completely new dependencies in the data. Finally, generation of rules from the tree by following the path from the root to leafs always leads to the condition redundancy which is often undesirable. 

\section{Methods}
\label{sec:methods}

\subsection{Basic notion}
Let $D(A, \delta)$ be the dataset of $|D|$ examples (observations, instances), each being characterized by a set of attributes $A=\{A_1, A_2,...,A_{|A|}\}$ and a label $\delta$. The meaning of the label depends on the problem. For classification tasks it corresponds to a discrete class identifier, i.e., $\delta \in \{L_1, L_2, ...,L_{|L|}\}$. In regression, it is a continuous value: $\delta \in \mathbb{R}$. In survival analysis, it represents the binary censoring status: $\delta \in \{0, 1\}$. In particular, the value of 0 indicates censored observations, also referred to as event-free (e.g., patients without disease recurrence), while 1 are non-censored examples, that were subject to an event (e.g., patients with recurrence). In the survival datasets, an additional variable $T$ representing the survival time, i.e. the time of the observation for event-free examples or the time before the occurrence of an event, must be specified.

The $i$th example of classification/regression dataset can be represented as a vector $x_i = (a_{i1}, a_{i2}, \ldots, a_{i|A|}, \delta_i)$; in survival problems it must be extended by the survival time: $x_i = (a_{i1}, a_{i2}, \ldots, a_{i|A|}, \delta_i, T_i)$. For simplicity, however, all types of datasets will be denoted as $D(A, \delta)$---the dependence of survival datasets on $T$ does not affect the idea of the presented algorithm.

Let $R$ be a set of rules generated by the induction algorithm, referred later as a rule-based data model or, simply, a model. Each rule $r \in R$ has the form:
$$
\IF c_1 \AND c_2 \AND \ldots \AND c_n \THEN \ldots
$$

The premise of a rule is a conjunction of conditions $c_j: A_j \odot a_j$, with $a_j$ being an element of the $A_j$ domain and $\odot$ representing a relation ($=$ for nominal attributes; $<, \leq, >, \geq$ for numerical ones). The conclusion of the rule can be a nominal value (classification), a numerical value (regression), or a Kaplan-Meier estimator~\cite{kaplan1958} of the survival function (survival analysis). Corresponding rules will be referred to as classification, regression, and survival rules, respectively. An example satisfying the conditions specified in the rule premise is stated to be covered by the rule.

Rule sets induced by our separate-and-conquer heuristic are unordered. Therefore, applying a model on the observation (e.g. a test example) requires evaluating set $R_\textrm{cov} \subseteq R$ of rules covering the example and aggregating the results. This differs from ordered rule sets (decision lists), where the first rule covering the investigated observation determines a model response. The method of aggregation depends on the problem. In classification, the output class label is obtained as a result of voting---each rule from $R_\textrm{cov}$ votes with its value of the quality measure used during the induction~\cite{wrobel2016}. In regression, the model response is an average of conclusions of $R_\textrm{cov}$ elements~\cite{sikora2012}. Similar situation is in the survival rules, but averaging concerns not numbers, but survival estimator functions~\cite{wrobel2017}.

\subsection{Separate-and-conquer}

The presented algorithm induces rules according to the separate-and-conquer principle \cite{furnkranz1999, michalski1973discovering}. Here we describe the fully automatic procedure---the user-guided variant is presented in the next subsection. An important factor determining performance and comprehensibility of the resulting model is a selection of a rule quality measure \cite{bruha1997, an2001rule, yao1999, wrobel2016} (rule learning heuristic \cite{furnkranz2005, janssen2010quest, minnaert}) that supervises the rule induction process. In the case of classification problems, our software provides user with a number of state-of-art measures calculated on the basis of the rule confusion matrix. Let $r$ be the considered classification rule. The examples whose labels are the same as the conclusion of $r$ will be referred to as positive, while the others will be called negative. The confusion matrix consists of the number of positive and negative examples in the entire training set ($P$ and $N$), and the number of positive and negative examples covered by the rule ($p$ and $n$). The idea can be straighyforwardly generalized for weighted examples by replacing numbers of examples in the confusion matrix by sums of their weights.
The measures built in the algorithm, e.g., C2 \cite{bruha1997}, Correlation \cite{furnkranz2005}, Lift \cite{bayardo}, RSS \cite{sikora2013data}, or s-Bayesian confirmation \cite{greco}, evaluate rules using various criteria resulting in very different models. For instance, RSS (also known as WRA \cite{furnkranz2005}) considers equally sensitivity ($p/P$) and specificity ($1 - n/N$) of the rule according to the formula $\textrm{RSS} = p/P - n/N$. Another common measure is conditional entropy which describes entropy of an outcome variable $Y$ given random variable $X$ as:
\begin{equation}
H(Y|X) = -\sum_{x \in X} P(x) \sum_{y \in Y} P(y|x) \log{P(y|x).}
\end{equation}
In our case $Y$ indicates class (positive/negative) and $X$ denotes whether rule covers the example (covered/uncovered). Therefore,
\begin{align}
 & P(X=\textrm{covered}) = \ (p + n)/(P + N), \\
 & P(Y=\textrm{positive}\ |\ X=\textrm{covered}) =\ p / (p + n), \\
 & P(Y=\textrm{positive}\ |\ X=\textrm{uncovered}) =\ (P - p) / (P + N - p - n).
 \end{align}
 The opposite probabilities, i.e., $P(X=\textrm{uncovered})$, $P(Y=\textrm{negative}\ |\ X=\textrm{covered})$, and $P(Y=\textrm{negative}\ |\ X=\textrm{uncovered})$ can be calculated straightforwardly by subtracting from 1 appropriate value.

The aforementioned measures are also used for evaluating regression rules, as regression is transformed by the algorithm to the binary classification problem. The transformation is done similarly as in~\cite{janssen2011}. Namely, the median $M$ and the standard deviation $\sigma$ of labels of instances covered by the rule $r$ is established. Observations from the entire set $D$ with labels from $[M-\sigma,M+\sigma]$ interval are assigned with a positive class. This allows determining elements of the confusion matrix and calculating all aforementioned quality measures. Note, however, that in contrast to classification problems, $P$ and $N$ values may change as rule coverage is modified.

The different situation is in the case of survival analysis, where rule outcomes are survival function estimates rather than numerical values. Thus, it is desirable for a rule to cover examples which survival distributions differ significantly from that of other instances. For this purpose, we use log-rank statistics \cite{harrington1982class} as a measure of survival rules quality. It is calculated as~$x^2/y$, where:
\begin{align}
x = & \sum\limits_{t \in T_c \cup T_u} (e^t_u - \frac{h^t_u}{h^t_c + h^t_u} \cdot (e^t_c + e^t_u)),\\
y = & \sum\limits_{t \in T_c \cup T_u} \frac{h^t_c \cdot h^t_u \cdot (e^t_c + e^t_u) \cdot (h^t_c + h^t_u - e^t_c - e^t_u)}{(h^t_c + h^t_u)^2 \cdot (h^t_c + h^t_u - 1)},
\end{align}
$T_c$ ($T_u$) is a set of event times of observations covered (uncovered) by the rule, $e_c^t$ ($e_u^t$) is the number of covered (uncovered) observations which experienced an event at time $t$, and $h_c^t$ ($h_u^t$) is the number of covered (uncovered) instances at hazard, i.e., still observable at time $t$.

\begin{algorithm}[!b]
	\small
	\begin{algorithmic}[1]
		\caption{Separate-and-conquer rule induction.}
		\label{alg:conquer}
		
		\Require
		$D(A,\delta)$---training dataset,
		\mincov---minimum number of yet uncovered examples that a new rule has to cover.
		\Ensure $R$---rule set.
		\State $D_{U} \gets D$	\Comment{set of uncovered examples}
		\State $R \gets \emptyset$ \Comment{start from an empty rule set}
		\Repeat
		\State $r \gets \emptyset$ \Comment{start from an empty premise}
		\State $r \gets \Call{Grow}{r, D, D_{U}, \mincov}$ \Comment{grow conditions}
		\State $r \gets \Call{Prune}{r,D}$ \Comment{prune conditions}
		\State $R \gets R \cup \{r\}$
		\State $D_{U} \gets D_{U}\setminus\Call{Cov}{r, D_U}$ \Comment{remove from $D_U$ examples covered by $r$}
		\Until{$D_{U} = \emptyset$}	
	\end{algorithmic}
\end{algorithm}

\begin{algorithm}[!t]
	\begin{algorithmic}[1]
		\caption{Growing a rule.}
		\label{alg:grow}
		\Require
		$r$---input rule,
		$D$---training dataset,
		$D_{U}$---set of uncovered examples,
		\mincov---minimum number of previously uncovered examples that a new rule has to cover.
		\Ensure
		$r$---grown rule.
		
		\Function{Grow}{$r$, $D$, $D_U$, $mincov$}
		
		\Repeat \Comment{iteratively add conditions}
		\State $c_\textrm{best} \gets \emptyset$ \Comment{current best condition}
		\State $q_\textrm{best} \gets -\infty,\quad \textrm{cov}_\textrm{best} \gets -\infty$ \Comment{best quality and coverage}
		
		\State $D_{r} \gets$ \Call{Cov}{$r$, $D$} \Comment{examples from $D$ satisfying $r$ premise}
		
		\For{$c \in$ \Call{GetPossibleConditions}{$D_r$}}
		\State $r_c \gets r \AND c$ \Comment{rule extended with condition $c$}
		\State $D_{r_c} \gets \Call{Cov}{r_c, D}$
		\If {$|D_{r_c} \cap D_U| \geq \mincov$} \Comment{verify coverage requirement}
		\State $q \gets$ \Call{Quality}{$D_{r_c}$, $D \setminus D_{r_c}$} \Comment{rule quality measure}
		
		\If {$q > q_\textrm{best}$ \textbf{or} ($q = q_\textrm{best}$ \textbf{and} $|D_{r_c}| > \textrm{cov}_\textrm{best}$)}
		\State $c_\textrm{best} \gets c,\quad q_\textrm{best} \gets q$,\quad $\textrm{cov}_\textrm{best} \gets |D_{r_c}|$
		\EndIf
		
		\EndIf
		\EndFor			
		\State $r \gets r \AND c_\textrm{best}$
		\Until{$c_\textrm{best} = \emptyset$}
		\State \Return{$r$}
		\EndFunction
	\end{algorithmic}
\end{algorithm}

Separate-and-conquer heuristic adds rules iteratively to the initially empty set as long as the entire dataset becomes covered (Algorithm~\ref{alg:conquer}). To ensure the convergence, every rule must cover at least $\mincov$ previously uncovered examples. The induction of a single rule consists of two stages: growing and pruning. In the former (presented in Algorithm~\ref{alg:grow}), elementary conditions are added to the initially empty premise. When extending the premise, the algorithm considers all possible conditions built upon all attributes (line 6: \textsc{GetPossibleConditions} function call), and selects those leading to the rule of highest quality (lines 10--12). In the case of nominal attributes, conditions in the form $A_j = a_j$ for all values $a_j$ from the attribute domain are considered. For continuous attributes, $A_j$ values that appear in the observations covered by the rule are sorted. Then, the possible split points $a_j$ are determined as arithmetic means of subsequent $A_j$ values and conditions $A_j < a_j$ and $A_j \geq a_j$ are evaluated. If several conditions render same results, the one covering more examples is chosen. Pruning can be considered as an opposite to growing. It iteratively removes conditions from the premise, each time making an elimination leading to the largest improvement in the rule quality. The procedure stops when no conditions can be deleted without decreasing the quality of the rule or when the rule contains only one condition. Finally, for comprehensibility, the rule is post-processed by merging conditions based on the same numerical attributes. E.g., conjunction $A_j \geq 3 \AND A_j \geq 5 \AND A_j < 10$ will be presented as $A_j \in [5,10)$.

In the regression and survival problems, the algorithm is performed once on the entire dataset $D$. For classification tasks, rules are induced independently for all classes. Particularly, when class $L_k$ is investigated, the set $D$ is binarized with respect to it: examples with labels equal to $L_k$ are positives, while the other are negatives. The detailed information about our algorithm for classification, regression, and survival rule induction using separate-and-conquer strategy can be found in~\cite{wrobel2016, wrobel2017}. The most important limitation of the presented approach is that induction is fully automatic---the user may control how the model looks like only by selecting quality measure and adjusting $\mincov$ parameter.

\subsection{Guided rule induction}
In order to allow user-guided rule induction, the separate-and-conquer heuristic explained in the previous subsection was extended. The preliminary step of the procedure is specifying user's requirements. They consists of several elements ordered by the priority (highest first):
\begin{enumerate}   \setlength{\itemsep}{0pt} \setlength{\parskip}{0pt} \setlength{\parsep}{0pt}
	\item $\Rexp$---set of initial (user-specified) rules which have to appear in the model. Depending on the parameters, initial rules are immutable or can be extended by other conditions (existing conditions cannot be altered, though).
	\item  $\Cp$/$\Ap$---multisets of preferred conditions/attributes. When deriving a rule, they are used before automatically induced conditions. The user may specify the multiplicity of each preferred element allowing it to be used in a given numbers of rules.
	\item $\Cm$/$\Am$---sets of forbidden conditions/attributes which cannot not appear in the automatically generated rules.
\end{enumerate}
In the classification problems, the requirements can be defined for each class separately. Additional parameters controlling guided rule induction are:
\begin{itemize}  \setlength{\itemsep}{0pt} \setlength{\parskip}{0pt} \setlength{\parsep}{0pt}
	\item $\Epref$/$\Eauto$---boolean indicating whether initial rules should be extended with a use of preferred/automatic conditions and attributes.
	\item $\Ypref$/$\Yauto$---boolean indicating whether new rules should be induced with a use of preferred/automatic conditions and attributes.
	\item $K_C$/$K_A$---maximum number of preferred conditions/attributes per rule.
	\item \textit{considerOtherClasses}---boolean indicating whether automatic induction should be performed for classes for which no user`s requirements have been defined (classification mode only).
\end{itemize}

\begin{algorithm}[!p]
	\small
	\begin{algorithmic}[1]
		\caption{Guided rule induction. Function \textsc{Grow} operate as in the fully automatic algorithm, but it excludes  attributes already present in $r$, forbidden attributes, and conditions intersecting with forbidden conditions.}
		\label{alg:conquer-expert}
		
		\Require
		$D(A,\delta)$---training dataset, $\Rexp$---set of initial rules,
		\Statex $\Cp$/$\Ap$---multiset of preferred conditions/attributes,
		\Statex $\Cm$/$\Am$---set of forbidden conditions/attributes,
		\Statex $K_C$/$K_A$---maximum number of preferred conditions/attributes per rule,
		\Statex $\Epref$/$\Eauto$---extend initial rules with preferred/automatic conditions (bool),
		\Statex $\Ypref$/$\Yauto$---induce new rules with preferred/automatic conditions (bool),
		\Statex \mincov---minimum number of yet uncovered examples that a new rule has to cover.
		
		\Ensure
		$R$---Rule set.
		
		\State $R \gets \emptyset$ \Comment start from empty rule set
		\State $D_{U} \gets D$	\Comment{set of uncovered examples}
		\For{$r \in \Rexp$}  \Comment{iterate over initial rules}
		
		\If{ $\Epref$} \Comment extend with preferred conditions/attributes
		\State $r \gets \Call{GuidedGrow}{r, D, D_U, \mincov, \Cp, \Ap, K_C, K_A}$
		\EndIf
		\If{$\Eauto$} \Comment extend with automatic conditions
		\State $r \gets \Call{Grow}{r, D, D_U, \mincov, \Cm, \Am}$ \Comment exclude forbidden knowledge
		\State $r \gets \Call{Prune}{r,D}$ \Comment{prune the rule}
		\EndIf
		\State $R \gets R \cup \{r\}$ \Comment add rule to rule set
		\State $D_{U} \gets D_{U}\setminus\Call{Cov}{r, D_U}$	
		\EndFor
		
		\If{$\Ypref$ \textbf{or} $\Yauto$} \Comment induce non-user rules
		\While{$D_{U} \neq \emptyset$}
		\State $r \gets \emptyset$ \Comment start from empty rule  		
		\If{$\Ypref$}
		\State $r \gets \Call{GuidedGrow}{r, D, D_U, \mincov, \Cp, \Ap, K_C, K_A}$
		\EndIf
		\If{$\Yauto$}
		\State $r \gets \Call{Grow}{r, D, D_U, \mincov, \Cm, \Am}$
		\State $r \gets \Call{Prune}{r,D}$ \Comment{prune the rule}
		\EndIf
		\State $R \gets R \cup \{r\}$ \Comment add rule to rule set
		\State $D_{U} \gets D_{U}\setminus\Call{Cov}{r, D_U}$	
		\EndWhile{}	
		\EndIf
		\State \Return{$R$}
	\end{algorithmic}
\end{algorithm}

The guided separate-and-conquer heuristic was presented in Algorithm~\ref{alg:conquer-expert}. It starts from processing initial rules in the order specified by the user. If $\Epref$ flag is enabled, an attempt is made to extend an initial rule by at most $K_C$ preferred conditions and $K_A$ preferred attributes (lines 4--5). After that, if $\Eauto$ flag is enabled, the algorithm adds automatically induced conditions using standard separate-and-conquer strategy (lines 6--9). When all initial rules have been processed, new ones are generated analogously; the corresponding boolean parameters are called $\Ypref$ and $\Yauto$ (lines 11--20).

For regression and survival problems, the described procedure is performed once, similarly as in the fully automatic mode. For classification tasks, the algorithm is executed for each class the knowledge has been specified for. If \textit{considerOtherClasses} parameter is set, this is followed by the fully automatic induction of rules for classes without user's preferences.

An important assumption concerning the semi-automatic induction is that knowledge elements are prioritized, i.e.:
\begin{itemize}  \setlength{\itemsep}{0pt} \setlength{\parskip}{0pt} \setlength{\parsep}{0pt}
	\item Initial rules and preferred conditions/attributes are more important than forbidden conditions/attributes. Therefore, if an initial rule $r \in \Rexp$ contains condition $c$ with attribute $A_j$, it will appear regardless of $A_j$ being marked as forbidden ($A_j \in \Am$) or $c$ intersecting one of the forbidden conditions ($\exists c' \in \Cm: c \AND c' \neq \emptyset$). The same holds for preferred conditions and attributes---forbidden knowledge applies to automatic induction only.
	
	\item Requirement of higher priority cannot be altered by that of lower priority. For instance, if an initial rule $r \in \Rexp$ contains condition $c$ with attribute $A_j$, $c$ cannot be modified, i.e., no other condition concerning $A_j$ can be added to this rule, neither preferred ($\Cp$ or $\Ap$-based), nor automatically induced. Similarly, preferred conditions cannot be overridden by \mbox{$\Ap$-based/automatic} conditions, etc.
	
	\item Requirements of the same category are prioritized by order in which they are specified by the user.
	
	\item User-defined knowledge cannot be subject to pruning.
\end{itemize}

\begin{algorithm}[!b]
	\small
	\begin{algorithmic}[1]
		\caption{Growing a rule using preferred conditions and attributes.  }
		\label{alg:grow-expert}
		
		\Require
		$r$---input rule, $D(A, \delta)$---training dataset,
		$D_{U}$---set of uncovered examples,
		\Statex $\Cp$/$\Ap$---multiset of preferred conditions/attributes,
		\Statex $K_C$/$K_A$---maximum number of preferred conditions/attributes per rule,
		\Statex \mincov---minimum number of yet uncovered examples that a new rule has to cover.
		
		\Ensure
		$r$---grown rule.
		
		\Function{GuidedGrow}{$r, D, D_U, \mincov, \Cp, \Ap, K_C, K_A$}
		\State $A_\textrm{free} \gets A \setminus \{ \Call{Attr}{r} \}$  \Comment exclude attributes already present in the rule
		
		\State $k_C \gets 0,\quad k_A \gets 0$ \Comment initialize counters with 0
		
		\Repeat \Comment{analyze preferred conditions}
		\State $c_\textrm{best} \gets \emptyset$ \Comment{current best condition}	
		
		\For{$c \in \Cp$}  \Comment{analyze all preferred conditions}
		\If{\Call{Attr}{$c$} $\in A_\textrm{free}$ \textbf{and} $|\Call{Cov}{r \AND c, D_U}| \geq \mincov$ \textbf{and}
			\par \hskip\algorithmicindent $\quad(r \AND c)$ \textrm{is better than} $(r \AND c_{best})$}
		\State $c_{best} \gets c$
		\EndIf
		\EndFor
		
		\State $r \gets r \AND c_{best}$	\Comment add condition to the rule premise
		\State $\Cp \gets \Cp \setminus \{c_{best}\}$ \Comment remove preferred condition
		\State $A_\textrm{free} \gets A_\textrm{free} \setminus \{\Call{Attr}{c_{best}}\}$ \Comment remove used attribute
		\State $k_C \gets k_C + 1$
		
		\Until{$k_C = K_C$ \textbf{or} $c_\textrm{best} = \emptyset$}
		
		\Repeat \Comment{analyze preferred attributes}
		\State $c_\textrm{best} \gets \emptyset$ \Comment{current best condition}	
		\For{$a \in \Ap \cap A_\textrm{free}$}  
		\State $c \gets \Call{InduceBestCondition}{a, D, D_U}$
		\If{ $|\Call{Cov}{r \AND c, D_U}| \geq \mincov$ \textbf{and} 
			\par \hskip\algorithmicindent $\quad(r \AND c)$ \textrm{is better than} $(r \AND c_{best})$}
		\State 	$c_{best} \gets c$
		\EndIf
		\EndFor
		\State $r \gets r \AND  c_{best}$	\Comment add condition to the rule premise
		\State $\Ap \gets \Ap \setminus \{a\}$ \Comment remove preferred attribute
		\State $A_\textrm{free} \gets A_\textrm{free} \setminus\{a\}$ \Comment remove used attribute
		\State $k_A \gets k_A + 1$
	
		\Until{$k_A = K_A$ \textbf{or} $c_\textrm{best} = \emptyset$}
		
		\State \Return{$r$}
		\EndFunction
	\end{algorithmic}
\end{algorithm}

The prioritization determines how a single rule is grown taking into account user's preferences (Algorithm~\ref{alg:grow-expert}). At the beginning, attributes already present in the rule are excluded (line 2). Then, at most $K_C$ preferred conditions fulfilling coverage requirement are added to the rule (lines 4--13). At each step, a condition rendering the rule of highest quality is selected (lines 6--8). After that, preferred attributes are processed similarly (lines 14--24). For each preferred attribute, a condition leading to the rule of highest quality is considered (line 17: \textsc{InduceBestCondition} function call). When a preferred condition/attribute is used, its multiplicity in $\Cp$/$\Ap$ multiset is decreased (lines 10, 21). Moreover, already employed attributes cannot be used again in the rule (lines 11, 22).

\section{Results}
\label{sec:results}
The algorithm was evaluated on three test cases representing classification, regression, and survival problems. The analysis of each dataset concerned:
\begin{enumerate}\setlength{\itemsep}{0pt} \setlength{\parskip}{0pt} \setlength{\parsep}{0pt}
\item {the validation of models rendered by automatic and user-guided rule induction; depending on the problem, this was done by 10-fold cross validation or train/test split},
\item {the analysis of rule sets induced on the entire datasets in the context of domain knowledge.}
\end{enumerate}
Table~\ref{tab:setting} presents problem-specific details of experimental procedures, e.g., model validation methods, quality criteria, statistical tests used for determining rules significance, etc.

The rule set descriptive statistics were common for all investigated datasets and consisted of: number of rules (\#rules), average number of conditions per rule (\#conditions), average rule precision ($p/(p+n)$) and support ($(p+n)/(P+N)$). Note, that the interpretation of indicators based on confusion matrix varies for different problems. Particularly, for classification $P$ and $N$ are fixed for each analyzed class, for regression $P$ and $N$ are determined for each rule on the basis of covered examples, for survival analysis all examples are considered positive, thus $N$ and $n$ equal to 0.

For all investigated models we report fraction of statistically significant rules at $\alpha=0.05$ level (\%significant). To control false discovery rate (FDR) in multiple testing, Benjamini-Hochberg correction was applied~\cite{benjamini1995}.

\begin{table}[!t]
	\setlength{\tabcolsep}{0.5em}
	\centering
	\caption{Experimental setting for investigated problems.}
	\label{tab:setting}
	\begin{small}
		\begin{tabular}{ccc}
			\hline	\hline
			\multicolumn{3}{c}{\textbf{Problem}} \\
			classification & regression & survival \\
			\hline
			\multicolumn{3}{c}{\textbf{Dataset}} \\
			\textit{seismic-bumps} & \ii{methane} & \ii{BMT-Ch}\\
			\hline
			\multicolumn{3}{c}{\textbf{Model validation method}} \\
			10-fold CV & train/test split & 10-fold CV \\
			\hline
			\multicolumn{3}{c}{\textbf{Quality criteria}} \\
			\parbox[t]{3.8cm}{\linespread{1.3}\centering
				sensitivity, specificity, and their geometric mean:
				$\mathrm{SE} = \frac{\mathrm{TP}}{\mathrm{TP}+\mathrm{FN}}$, $\mathrm{SP} = \frac{\mathrm{TN}}{\mathrm{TN}+\mathrm{FP}}$, \\
				$\mathrm{Gm} = \sqrt{\mathrm{SE} \times \mathrm{SP}}$
			} &
			\parbox[t]{3.8cm}{\linespread{1.0}\centering
				root relative \\ squared error \\
				$\textrm{RRSE} = \sqrt{\frac{\sum_{i=1}^{|D|}(\widehat{\delta_i} - \delta_i)^2}{\sum_{i=1}^{|D|}(\overline{\delta} - \delta_i)^2} }$ \\
				$\delta_i$---observed label, $\widehat{\delta_i}$---expected, $\overline{\delta}$---average} &
			\parbox[t]{3.8cm}{\linespread{1.3}\centering
				integrated Brier score $\mathrm{IBS}$\\ (see~\cite{wrobel2017} for details)}\\
			\hline
			\multicolumn{3}{c}{\textbf{Quality difference significance test}} \\
			Student's $t$-test & --- & Student's $t$-test\\
			\hline
			\multicolumn{3}{c}{\textbf{Rule significance test}} \\
			\parbox[t]{3.8cm}{Fisher's exact test for comparing confusion matrices} &
			\parbox[t]{3.8cm}{$\chi^2$ test for comparing label variance of covered vs. uncovered examples} &
			\parbox[t]{3.8cm}{log-rank for comparing survival functions of covered vs. uncovered examples}\\
			\hline	\hline
		\end{tabular}
	\end{small}
\end{table}

Another analysis step was the comparison of similarity between guided-guided and automatic rule sets. The similarity between two rule sets $R_1$ and $R_2$ on the dataset $D$ is expressed as:
\begin{equation}
\mathrm{similarity} = (a+b) \Big/ {\binom{|D|}{2}}, \textrm{where}
\end{equation}

\begin{itemize} \setlength{\itemsep}{0pt} \setlength{\parskip}{0pt} \setlength{\parsep}{0pt}
	\item $a$ is the number of pairs $\{e_1, e_2\}$ of examples in $D$ for which there exists some rule in $R_1$ and some rule in $R_2$ covering both $e_1$ and $e_2$.
	
	\item $b$ is the number of pairs $\{e_1, e_2\}$ of examples in $D$ for which there neither exists rule in $R_1$ nor in $R_2$ covering both $e_1$ and $e_2$.
	
	\item $\binom{|D|}{2}$ is the number of all pairs of examples in $D$.
\end{itemize}

The measure might be interpreted as the probability of agreement between two rule sets for randomly chosen pair of examples. The agreement between rule sets $R_1$ and $R_2$ for pair of examples $\{e_1, e_2\} \in D$ means that:

\begin{itemize} \setlength{\itemsep}{0pt} \setlength{\parskip}{0pt} \setlength{\parsep}{0pt}
	\item if both examples $\{e_1, e_2\}$ satisfy premise of some rule in $R_1$ then they also both satisfy premise of some rule in $R_2$,
	
	\item if examples $\{e_1, e_2\}$ are not covered by common rule in $R_1$ then they also are not covered by common rule in $R_2$,
	
	\item if both examples $\{e_1, e_2\}$ are not covered by any of rules in $R_1$ then they also are not covered by any of rules in $R_2$,
	
	\item if one of examples $\{e_1, e_2\}$ is covered by some rule in $R_1$ and the other one is not covered by any of rules in $R_1$ then the same applies to $R_2$.
	
\end{itemize}

The rule sets similarity measure takes values between 0 and 1. The value 0 indicates that the two rule sets do not agree of any pairs of examples from given dataset $D$. The value 1 means the perfect agreement, i.e. that there not exists any pair of examples $\{e_1, e_2\}$ which are covered by common rule in one of the rule sets and not covered by common rule in the second one. Since proposed measure evaluates the similarity between subsets of examples covered by rule sets, it is not influenced by rules overlap within a set. In particular, if rule sets $R_1$ and $R_2$ have similarity score equal 1, then extending these rule sets by additional rules does not change the value of the score.

The proposed similarity score can be also considered as a variant of Rand measure \cite{rand1971objective} used for evaluation of clustering performance. However, our proposal takes into account that single example can satisfy premises of several rules as well as that it may be not covered by any of the rules.

The following subsections contain detailed analysis of classification, regression, and survival experiments.

\subsection{Classification}
Classification experiments were performed on \textit{seismic-bumps} dataset from UCI Machine Learning Repository \cite{uci}. The dataset had been prepared and made available by the authors of the paper and concerns a problem of forecasting high energy seismic bumps in coal mines \cite{kabiesz}. It contains 2\,584 instances (170 positives and 2\,414 negatives) and 19 attributes characterizing seismic activity in the rock mass within one 8-hour shift (see Table~\ref{tab:seismic} for description of crucial features). The value 1 of \textit{class} attribute indicates the presence of a seismic bump with energy higher than $10^{4}$ J in the next shift.

\begin{table}[!t]
	\linespread{1.0}
	\setlength{\tabcolsep}{0.5em}
	\centering
	\caption{Description of selected attributes of \textit{seismic-bumps} dataset.}
	\label{tab:seismic}
		\begin{small}
	\begin{tabular}{p{2.2cm}p{9.3cm}}
		\hline
		attribute & description \\
		\hline
		\textit{seismic}\newline (\textit{seismoacoustic}) & result of shift seismic (seismoacoustic) hazard assessment in the mine working obtained by the seismic (seismoacoustic) method developed by mine experts (a---lack of hazard, b---low hazard, c---high hazard, d---danger state)\\
		
		\textit{genergy} & seismic energy recorded within the previous shift by the most active geophone (GMax) out of geophones monitoring the longwall\\
		
		\textit{gimpuls} & a number of pulses recorded within the previous shift by GMax\\
		
		\textit{goenergy} & a deviation of energy recorded within the previous shift by GMax from the average energy recorded during eight previous shifts\\
		
		\textit{goimpuls} & a deviation of a number of GMax pulses within the previous shift from the average number of pulses within eight previous shifts\\
		
		\textit{ghazard} & result of shift seismic hazard assessment in the mine working obtained by the seismoacoustic method based on registration coming from GMax only\\
		
		\textit{nbumpsX} & a number of seismic bumps in the energy range $[10^X,10^{(X+1)})$ (where $X \in \{2,3,..,8\}$), registered within the previous shift\\
		
		\textit{senergy} & total energy of seismic bumps registered within the previous shift\\
		
		\textit{maxenergy} & the maximum energy of the seismic bumps registered within the previous shift\\
		
		\hline
	\end{tabular}
\end{small}
\end{table}

The model validation was carried out according to the stratified 10-fold cross validation. To establish algorithm parameters, automatic rule induction was performed as an initial step. Due to strong imbalance of the problem, a geometric mean (Gm) of sensitivity and specificity was used for assessment. Among examined quality measures (C2, Correlation, Conditional entropy, Lift, RSS, SBayesian), Conditional entropy with $\mincov=11$ was selected for further investigation.

To demonstrate the flexibility of our algorithm, a guided rule induction was done in several variants, with different algorithm parameters.
The variants marked as guided-c1, guided-c2, guided-c3, guided-c4 are attempts to use in the classifier attributes that, according to the domain knowledge, should have the greatest significance for bumps forecasting \cite{kabiesz}. The guided-c5 and guided-c6 variants are an attempt to define the classifier only on the basis of data coming from one measurement system: in the former it is forbidden to use attributes containing data from geophones (i.e., a seismoacoustic system), in the latter it is forbidden to use attributes containing data from seismometers (i.e., a seismic system).

The variants together with corresponding algorithm parameters are listed below. Class-specific requirements are defined with superscripts, e.g., $\Ap^0$ contains preferred attributes for class 0 (lack of superscript indicates that knowledge applies to both classes). Only important parameters are specified.
\begin{description}
\item[\textbf{guided-c1:}] Model consists of two initial rules:\\
{\small
	$\Rexp = \{
	 	(\IF \ii{gimpuls} < 750 \THEN \ii{class} = 0),
	 	(\IF \ii{gimpuls} \geq 750 \THEN \ii{class} = 1)
	 	\},
	\Cp = \Cm = \Ap = \Am = \phi, \Epref = \Eauto = \Ypref = \Yauto = \textbf{false}$.
}
\item[\textbf{guided-c2:}] Attribute \ii{gimpuls} is used in rules for both classes at least once:\\
{\small
	$\Ap = \{gimplus^1\},
	\Rexp = \Cp = \Cm = \Am = \phi, \Ypref = \Yauto = \textbf{true}, K_A = 1$.
}
\item[\textbf{guided-c3:}] At least $2/3$ of rules contain \ii{gimpuls}, \ii{genergy}, and \ii{senergy} attributes together:\\
{\small
	$\Ap = \{gimplus^{100}, genergy^{100}, senergy^{100}\},
	\Rexp  = \Cp = \Cm = \Am = \phi,  \Ypref = \Yauto = \textbf{true}$.
}
\item[\textbf{guided-c4:}] At least one of \ii{seismic}, \ii{seismoacoustic}, and \ii{ghazard} attributes is used in each rule, with an additional requirement on value sets---class 0 may use values \ii{a, b}, class 1 may use values \ii{b, c, d}:\\
{\small
	$\Cp^0 = \{
		(\ii{seismic} = \ii{a})^\infty, (\ii{seismic} = \ii{b})^\infty,
		(\ii{seismoacoustic} = \ii{a})^\infty, (\ii{seismoacoustic} = \ii{b})^\infty,
		(\ii{ghazard} = \ii{a})^\infty, (\ii{ghazard} = \ii{b})^\infty \},
	\Cp^1 = \{
		(\ii{seismic} = \ii{b})^\infty, (\ii{seismic} = \ii{c})^\infty,
		(\ii{seismic} = \ii{d})^\infty,
		(\ii{seismoacoustic} = \ii{b})^\infty, (\ii{seismoacoustic} = \ii{c})^\infty, (\ii{seismoacoustic} = \ii{d})^\infty,
		(\ii{ghazard} = \ii{b})^\infty, (\ii{ghazard} = \ii{c})^\infty, (\ii{ghazard} = \ii{d})^\infty \},\\
	\Rexp  = \Cm = \Ap = \Am = \phi, \Ypref = \Yauto = \textbf{true}, K_C = 1$.
}
\item[\textbf{guided-c5:}] Attributes \ii{gimpuls}, \ii{goimpuls}, \ii{ghazard}, and \ii{seismoacoustic} are forbidden:\\
{\small
	$\Am = \{\ii{gimpuls}, \ii{goimpuls}, \ii{ghazard}, \ii{seismoacoustic}\},\\
	\Rexp = \Cp = \Cm = \Ap = \phi, \Yauto = \textbf{true}$.
}
\item[\textbf{guided-c6:}] Attributes from \ii{nbumps} family as well as \ii{senergy}, \ii{maxenergy}, and \ii{seismic} are forbidden: analogous to guided-c5.
\end{description}

\begin{table}[!b]
	\setlength{\tabcolsep}{0.30em}
	\centering
	\caption{The analysis of the classification rule sets in terms of model quality (SE---sensitivity, SP---specificity, Gm---their geometric mean, Gm-$p$---Student's t-test $p$-value comparing Gm of user's variants w.r.t. auto) and descriptive statistics.}
	\label{tab:classification}
	\begin{tabular}{lccccccccccccc}
		\hline
		&& \multicolumn{4}{c}{Validation} &&  \multicolumn{6}{c}{Descriptive statistics}&\\
		&& \multicolumn{4}{c}{(10-CV)} &&  \multicolumn{6}{c}{(full dataset)}\\
		\cline{3-6}  \cline{8-14}
		variant && SE & SP & Gm   & Gm-$p$ && \rot{\#rules} & \rot{\#conditions} & \rot{support} & \rot{precision} & \rot{\%significant} & \rot{similarity}\\
		\hline
		auto && 0.67 & 0.76 & 0.708 $\pm$ 0.071 & \e\e--- 		&& \e67 & 7.2 &  0.14 & 0.51 & \e94 & ---\\
		guided-c1 && 0.49 & 0.813 & 0.627 $\pm$ 0.064 & \e\e0.01 	&& \e\e2 & 1.0 & 0.50 & 0.55 & 100 & 0.74\\
		guided-c2 && 0.62 & 0.82 & 0.711 $\pm$ 0.062 & \e\e0.90 	&& \e39 & 5.0 & 0.30 & 0.57 & \e95 & 0.92\\
		guided-c3 && 0.58 & 0.82 & 0.690 $\pm$ 0.046 & \e\e0.42		&& 155 & 5.0 & 0.38 & 0.78 & \e97 & 0.88\\
		guided-c4 && 0.42 & 0.81 & 0.580 $\pm$ 0.057 & $< 0.01$	&& 121	& 5.9 & 0.31 & 0.80 & \e99 & 0.51\\
		guided-c5 && 0.64 & 0.78 & 0.701 $\pm$ 0.082 & \e\e0.65		&& \e43 & 4.1 & 0.30 & 0.48 & \e93 & 0.88\\
		guided-c6 && 0.56 & 0.73 & 0.622 $\pm$ 0.070 & \e\e0.02		&& \e55 & 4.9 & 0.20 & 0.49 & \e96 & 0.88\\
		\hline
	\end{tabular}
\end{table}

The summary of results for automatic and guided classification variants are given in Table \ref{tab:classification}. Below there is also an analysis of the rule sets obtained by means of the automatic, guided-c1, guided-c2, guided-c4, and guided-c6 methods on the entire dataset.

The rule set induced automatically consisted of 67 rules with average support and precision equal to 0.14 and 0.51, respectively (taking into account dataset imbalance, it is an acceptible result). The attributes \textit{goimplus}, \textit{gimplus}, \textit{ghazard}, and \textit{seismoacoustic} occured in 49, 47, 18, and 13 rules, respectively.

Below we present –- for each decision class –- the strongest rule generated automatically. In brackets there are confusion matrix elements which allow calculating support and precision. The rules indicating decision class 1 were more specific, less precise, and covered less examples.

\begin{description} \setlength\itemsep{0pt} \setlength{\parskip}{0pt} \setlength{\parsep}{0pt}
{\small
	\item[]
		$\IF \ii{gimpuls} < 218 \AND \ii{goimpuls} < -1.5 \AND \ii{nbumps} < 2 \THEN \ii{class} = 0$\\
		($p=565$, $n=6$, $P=2414$, $N=170$)
	\item{}
		$\IF \ii{goenergy} < 96 \AND \ii{maxenergy} \geq 1500 \AND \ii{gimpuls} \in (541, 2258) \AND \ii{goimpuls} \in (-34, 95) \AND \ii{genergy} \in (61250, 662435) \AND \ii{senergy} < 36050 \AND \ii{nbumps3} <5 \AND \ii{nbumps} > 1 \THEN \ii{class} = 1$ ($p=29$, $n=19$, $P=170$, $N=2414$)
}
\end{description}

\noindent
The characteristics of rules obtained in the guided-c1 experiment are as follows:

\begin{description} \setlength\itemsep{0pt} \setlength{\parskip}{0pt}
{\small
	\item[]
	$\IF \ii{gimpuls} < 750 \THEN \ii{class} = 0$ ($p=1962$, $n=89$, $P=2414$, $N=170$)
	\item[]
	$\IF \ii{gimpuls} \geq 750 \THEN \ii{class} = 1$ ($p=81$, $n=452$, $P=170$, $N=2414$)
}
\end{description}

\noindent
The classifier based on these two rules had a significantly worse classification quality. However, it is worth noticing that the first rule was less precise by only 3.3\% than the best rule generated automatically for this class, however, its support was 3.6 times larger. The rule pointing at class 1 had the precision of 0.15 which was over twice as much as the 0.065 \textit{a priori} precision of this class.

The guided-c2 experiment aimed at forcing the occurrence of the \textit{gimpuls} attribute in each rule as well as adding other elementary conditions to the rule premise. As it can be seen in Table~\ref{tab:classification}, this leaded to a model with the best classification ability. In addition, the number of rules decreased, compared to the automatically generated rule set, while their average support and average precision increased over 214\% and 11\%, respectively.

The results achieved for the guided-c6 experiment show that it is impossible to obtain a good quality classifier only on the basis of data coming from a seismic system, thus it is indispensable to use geophones (sensors which register seismoacoustic emission).

In all cases a majority of induced rules were statistically significant. Rule sets generated under the guided mode (particularly guided-c2 and guided-c3 variants) were less numerous than those generated automatically. They also contained fewer elementary conditions. According to the value of similarity measure rule sets induced in auto, guided-c2 and guided-3 mode were very similar. However, rules generated in the guided mode represented knowledge which is more in compliance with the user’s requirements and intuition. Additionally, the analysis of standard deviations shows that rule sets generated in the guided mode were more stable in their classification abilities.

The experiments we carried out show that the guided (interactive) model definition allows verifying certain research hypotheses and, in particular, obtaining classifiers superior to those generated automatically. The induction of successive rule sets may contribute to further analyses. For example, one could attempt to develop a classifier made of the first rule from guided-c1 model supplemented with automatic rules. Our software enables performing many variants of such analyses.

\subsection{Regression}
\begin{figure}[h!]
	\centering
	\includegraphics[width=0.5\textwidth]{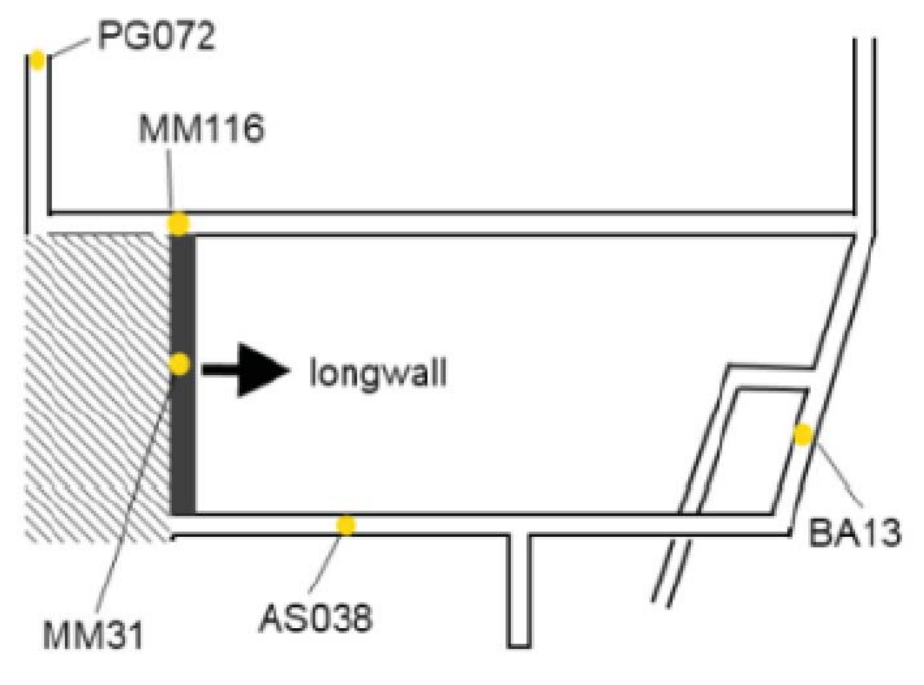}
	\caption{Sensor location in the longwall area.
		\label{fig:me}}
\end{figure}

The usability of the presented algorithm for regression problems was verified on \ii{methane} dataset, which concerns the problem of predicting methane concentration in a coal mine. The set contains 13\,368 train and 5\,728 test instances characterized by 7 attributes. The features indicate methane concentration (\ii{MM116}, \ii{MM31} [\%]), air velocity (\ii{AS038} [m/s]), airflow (\ii{PG072} [m/s]), atmospheric pressure (\ii{BA13} [hPa]), and whether the coal production process ($\ii{PD}=1$) is carried out. The location  of sensors is depicted in Fig. \ref{fig:me}. The attributes represents averaged measurements from 30-second periods. The task is to predict the maximal value of methane concentration registered by \ii{MM116} for next 3 minutes.

As in the previous case, an automatic induction was done in order to adjust parameters. Eventually, RSS quality measure with $\mincov=4$ was selected as providing the best trade off between root relative squared error (RRSE) and model complexity expressed by the number of rules and conditions. The following variants of user's knowledge were investigated:
\begin{description}
\item[\textbf{guided-r1:}] The model contains $\ii{PD} = 0$ and $\ii{PD} = 1$ conditions, both appearing in three rules:\\
{\small
	$\Cp = \{ (\ii{PD} = 0)^3, (\ii{PD} = 1)^3 \},\\
	\Rexp = \Cm = \Ap = \Am = \phi, \Ypref = \Yauto = \textbf{true}, K_C = 1$.
}
\item[\textbf{guided-r2:}] The conjunction $\ii{PD} = 1 \AND \ii{MM116} < 1$ appears in five rules:\\
{\small
	$\Cp = \{(\ii{PD} = 1 \AND \ii{MM116} < 1)^5\},\\
	\Rexp = \Cm = \Ap = \Am = \phi,  \Ypref = \Yauto = \textbf{true}, K_C = 1$.
}
\item[\textbf{guided-r3:}] The conjunction $\ii{PD} = 0 \AND \ii{MM116} > 1$ appears in five rules: analogous to guided-r2.
\item[\textbf{guided-r4:}] Attributes \ii{DMM116}, \ii{MM116}, and \ii{PD} appear in every rule:\\
{\small
	$\Ap = \{\ii{DMM116}^\infty, \ii{MM116}^\infty, \ii{PD}^\infty\},\\
	\Rexp = \Cp = \Cm = \Am = \phi, \Ypref = \Yauto = \textbf{true}, K_A = 3$.
}
\end{description}

\begin{table}[!b]
	\setlength{\tabcolsep}{0.5em}
	\centering
	\caption{The analysis of the regression rule sets in terms of model quality (RRSE---relative root squared error) and descriptive statistics.}
	\label{tab:regression}
	\begin{tabular}{lcccccccccc}
		\hline
		&& \multicolumn{1}{c}{Validation} &&  \multicolumn{6}{c}{Descriptive statistics}&\\
		&& \multicolumn{1}{c}{(train/test)} &&  \multicolumn{6}{c}{(entire dataset)}\\
		\cline{3-3}  \cline{5-11}
		variant && RRSE 		&& \rot{\#rules} & \rot{\#conditions} & \rot{support} & \rot{precision} & \rot{\%significant} & \rot{similarity}\\
		\hline
		auto && 0.918 			&& \e9 & 3.5 & 0.26 & 0.64 & \e88 &---\\
		guided-r1 && 0.811 		&& 19 & 4.4 & 0.17 & 0.66 & \e95 & 0.70\\
		guided-r2 && 0.793 		&& 11 & 3.3 & 0.18 & 0.69 & 100 & 0.93\\
		guided-r3 && 0.863 		&& \e8 & 2.9 & 0.18 & 0.78 & \e87 & 0.93\\
		guided-r4 && 1.174 		&& 41 & 5.5 & 0.10 & 0.70 & 100 & 0.60\\
		\hline
	\end{tabular}
\end{table}

The automatic induction produced 9 rules, which allowed achieving RRSE of 0.918, i.e. smaller than the naive prognosis based on the average value of the dependent variable (see Table~\ref{tab:regression} for all the results). The \ii{MM116} and \ii{MM31} attributes dominated in the rule premises. This means that the currently registered concentration of methane has the largest impact on the future concentration.  This is illustrated, for example, by the following rule:

\begin{description} \setlength\itemsep{0pt} \setlength{\parskip}{0pt}
{\small
	\item[]
	$\IF \ii{MM31} < 0.225 \THEN \ii{MM116Pred} = 0.4\ [0.39, 0.41]$
}
\end{description}
\noindent
which shows that if the concentration of methane in the middle of the longwall is low, the predicted concentration at the longwall exit will be about twice as high (it will remain in the range [0.39, 0.41]).

Another rule presents an interesting dependence. If the methane concentration is on an average level (about 1\%), too high air velocity can lead to the eddies of the gas mixture at the longwall exit and, at the same time, can increase the methane concentration (methane in the range [0.92, 1.28])

\begin{description} \setlength\itemsep{0pt} \setlength{\parskip}{0pt}
{\small
	\item[]
	$\IF \ii{MM116} \geq 0.85 \AND \ii{AS038} \geq 2.05 \THEN \ii{MM116PRed} = 1.1\  [0.92, 1.28]$
}
\end{description}

In automatically generated rules, the PD attribute occured only twice. Within the guided-r1 experiment the use of elementary conditions $\ii{PD}=1$ or $\ii{PD}=0$ (the cutter-loader is not working) was obligatory. This reflects the hypothesis supported by domain knowledge that the emission of methane is larger while the cutter-loader is working. In this way, a significant error reduction was achieved at the cost of increasing the number of rules and conditions occurring in their premises.

Within the guided-r3 experiment, the occurrence of the $\ii{PD}=1$ and $\ii{MM116}<1$ conditions at the same time was obligatory. A rule containing only those conditions covered 14\% of all examples and looks as follows:

\begin{description} \setlength\itemsep{0pt} \setlength{\parskip}{0pt}
{\small
	\item[]
	$\IF \ii{PD} = 1 \AND \ii{MM116} < 1 \THEN \ii{MM116PRed} = 0.8\ [0.63, 0.97]$
}
\end{description}
\noindent
The induction of the above rule allowed better identification of rules indicating higher methane concentration, e.g.:

\begin{description} \setlength\itemsep{0pt} \setlength{\parskip}{0pt}
{\small
	\item[]
	$\IF \ii{PD} = 1 \AND \ii{MM116} \geq 0.95 \AND \ii{AS038} \geq 2.25 \AND \ii{PG072} \in (1.75, 1.95) \THEN\\ \ii{MM116PRed} = 1.5\ [1.33, 1.67]$
}
\end{description}
\noindent
and, as a result of that, caused further decrease of RRSE.

Apart from the last case (guided-r4), rule sets induced in guided mode produced a smaller RRSE values than an automatically generated set. The guided-r1 settings enforce the use of \ii{PD} = 0 and \ii{PD} = 1 conditions in three rules. This algorithm settings reflect an attempt to make the methane level dependent on the coal production process. The regression errors of guided-1 and automatically generated were close to each other, while the value of the similarity measure was relatively low. This means that both of these rule sets generated different coverages of the example space.

Generally, in the case of regression rule induction, the definition of the user’s requirement and the analysis of the rules can be difficult because there are no explicitly defined decision classes here. However, as we can see, an interactive analysis allows reducing estimation error. In addition, it is possible to identify interesting regularities in the data, similar to the negative effects of too high air velocity.

\subsection{Survival analysis}
Another area our algorithm can be applied is survival analysis. The corresponding experiments were performed on \ii{BMT-Ch} dataset, which describes 187 pediatric patients (75 females and 112 males) with several hematologic diseases: 155 malignant disorders (i.a. 67 patients with acute lymphoblastic leukemia, 33 with acute myelogenous leukemia, 25 with chronic myelogenous leukemia, 18 with myelodysplastic syndrome) and 32 nonmalignant cases (i.a. 13 patients with severe aplastic anemia, 5 with Fanconi anemia, 4 with X-linked adrenoleukodystrophy). All patients were subject to the unmanipulated allogeneic unrelated donor hematopoietic stem cell transplantation. Instances are described by 37 conditional attributes, the meaning of the selected ones is as follows: \ii{relapse}---reoccurrence of the disease, \ii{PLTRecovery}---time to platelet recovery defined as platelet count $>50000/mm3$, \ii{ANCRecovery}---time to neutrophils recovery defined as neutrophils count $>0.5 x 10^{9}/L$, \ii{aGvHD\_III\_IV}---development of acute graft versus host disease stage III or IV, \ii{extcGvHD}---development of extensive chronic graft versus host disease, \ii{CD34} $(10^{6}/kg)$---\ii{CD34+} cell dose per kg of recipient body weight, \ii{CD3} $(10^{8}/kg)$---\ii{CD3+} cell dose per kg of recipient body weight. Patient's death is considered as an event in the survival analysis.

The remaining attributes concern coexisting diseases/infections (e.g. cytomegalic inclusion disease) and describe matching between the bone marrow donor and recipient.

The experiments were performed for $\mincov=5$ with different initial knowledge variants (note, that in the survival analysis class labels for initial rules cannot be specified):
\begin{description}
\item[\textbf{guided-s1:}] Every rule contains \ii{CD34} and does not contain \ii{ANCRecovery} and \ii{PLTRecovery} attributes:\\
{\small
	$\Ap = \{\ii{CD34}^\infty\}, \Am =  \{\ii{ANCRecovery}, \ii{PLTRecovery}\},\\
	\Rexp = \Cp = \Cm = \phi, \Ypref = \Yauto = \textbf{true}, K_A = 1$.
}
\item[\textbf{guided-s2:}] The model consists of four expert rules:\\
{\small
	$\Rexp = \{
		(\IF \ii{extcGvHD} = \ii{No} \AND \ii{CD34} < 10 \THEN \ldots),\\
		(\IF \ii{extcGvHD} = \ii{No} \AND \ii{CD34} \geq 10 \THEN \ldots),\\
		(\IF \ii{extcGvHD} = \ii{Yes} \AND \ii{CD34} < 10 \THEN \ldots),\\
		(\IF \ii{extcGvHD} = \ii{Yes} \AND \ii{CD34} \geq 10 \THEN \ldots)
	\},\\
	\Cp = \Cm = \Ap = \Am = \phi,  \Epref = \Eauto = \Ypref = \Yauto = \textbf{false}$.
}
\item[\textbf{guided-s3:}] Similarly as in the previous case, but \ii{CD34} ranges may be altered and rules can be extended with automatic conditions:\\
{\small
	$\Rexp = \{
	(\IF \ii{extcGvHD} = \ii{No} \THEN \ldots)^2, (\IF \ii{extcGvHD} = \ii{Yes} \THEN \ldots)^2\},\\
	\Ap = \{ \ii{CD34}^\infty\},\\
	\Cp = \Cm = \Am = \phi,  \Epref = \Eauto = \textbf{true}, \Ypref = \Yauto = \textbf{false}, K_A = 1$.
}
\item[\textbf{guided-s4:}] The model consists of two initial rules:\\
{\small
	$\Rexp = \{ (\IF \ii{CD34} < 10 \THEN \ldots), (\IF \ii{CD34} \geq 10 \THEN \ldots)\},\\
	\Cp = \Cm = \Ap = \Am = \phi,  \Epref = \Eauto = \Ypref = \Yauto = \textbf{false}$.
}
\end{description}

\begin{table}[!b]
	\setlength{\tabcolsep}{0.4em}
	\centering
	\caption{The analysis of the survival rule sets in terms of model quality (IBS---integrated Brier score, IBS-$p$---Student's t-test $p$-value comparing IBS of user variants w.r.t. auto) and descriptive statistics.}
	\label{tab:survival}
	\begin{tabular}{lcccccccccccc}
		\hline
		&& \multicolumn{2}{c}{Validation} &&  \multicolumn{6}{c}{Descriptive statistics}&\\
		&& \multicolumn{2}{c}{(10-CV)} &&  \multicolumn{6}{c}{(entire dataset)}\\
		\cline{3-4}  \cline{6-12}
		variant && IBS & IBS-$p$ 		&& \rot{\#rules} & \rot{\#conditions} & \rot{support} & \rot{precision} & \rot{\%significant} & \rot{similarity}\\
		\hline
		auto 		&& 0.212 $\pm$ 0.048	& --- 	&& \e4 & 3.0 & 0.49 & 1.00 & 100 &---\\
		guided-s1 	&& 0.235 $\pm$ 0.069 & 0.31	&& 14 & 4.1 & 0.14 & 1.00 & \e71 & 0.30\\
		guided-s2 	&& 0.221 $\pm$ 0.033 & 0.48	&& \e4 & 2.0 & 0.21 & 1.00 & \e50 & 0.27\\
		guided-s3		&& 0.225 $\pm$ 0.036 & 0.43	&& \e4 & 3.0 & 0.36 & 1.00 & 100 & 0.49\\
		guided-s4 	&& 0.223 $\pm$ 0.026 & 0.38	&& \e2 & 2.0 & 0.50 & 1.00 & 100 & 0.48\\
		\hline
	\end{tabular}
\end{table}

Detailed results can be found in Table~\ref{tab:survival}.
The automatic method generated four rules in which the survival function depended on such factors as the patient age, donor age, gender match, disease relapse, and the number of days to achieve the PLTRecovery. The variants of the guided rule induction refer to the verification of the research hypothesis that an increased dosage of \ii{CD34+} cells/kg extends overall survival time without simultaneous occurrence of undesirable events affecting the patients' quality of life.

The guided-s2 experiment was based on an arbitrary definition of rules. These rules try to make the survival function dependent on the \ii{CD34} dosage and the occurence of the extensive chronic graft versus host disease. The average $p$-value of the rules with FDR correction was 0.229, which shows that the rules considered separately did not contain statistically useful information. As it can be observed, the IBS value was also worse. Better results were achieved for the rules containing only the \ii{CD34} attribute (guided-s4). They were characterized by the average $p$-value of 0.019 after correction.

In the guided-s3 experiment the use of the \ii{CD34} and \ii{extcGvHD} attributes was obligatory, but in the former case, the division point on the attribute domain was not defined. It was also admissible to add other attributes to the rule premises. The algorithm generated four rules with the average $p$-value after correction equaled to 0.12.

Figure \ref{fig:scurve} shows survival curves corresponding to the three rules presented below. Survival curve for the entire set of examples is also given.

\begin{figure}[b!]
	\centering
	\includegraphics[width=\textwidth]{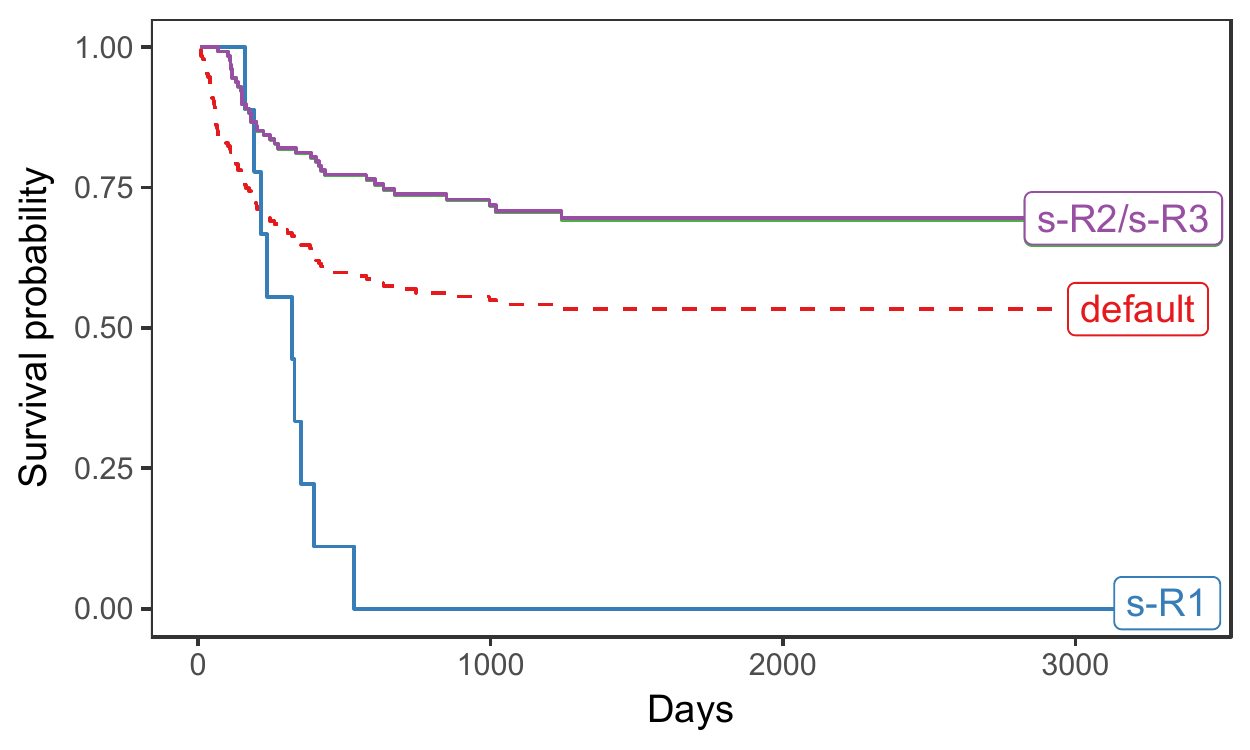}
	\caption{Survival curves of observations covered by three selected rules and the entire set of examples (default).
		\label{fig:scurve}}
\end{figure}

\begin{description} \setlength\itemsep{0pt} \setlength{\parskip}{0pt}
{\small
	\item[s-R1]
	$\IF \ii{CD34} \leq 13.055 \AND \ii{extcGvHD} = Yes \AND \ii{StemCellSrc} = Peripheral\_blood \AND \ii{ANCRecovery} \leq 15$
	\item[s-R2]
	$\IF \ii{extcGvHD} = No \AND \ii{CD34} \geq 0.8$
	\item[s-R3]
	$\IF \ii{extcGvHD} = No$
}
\end{description}
\noindent
There was practically no difference between survival curves corresponding to the second and the third rule. One can see that the second rule was more specific then the third one. The \ii{CD34} dosage does not have any impact on the survival function if the patient has not extensive chronic graft versus host disease.

According to the medical knowledge, chronic graft versus host disease remains dangerous complication of allogeneic stem cell transplantation. However, mild forms of this disease are often manageable and if the disease is under control, extend the overall survival time as it causes the elimination of cancer cells (blasts) remaining in the blood. On the other hand, the first rule shows that the patients with small doses of \ii{CD34} who developed extensive chronic graft versus host disease have significantly shorter survival time in spite of fast neutrophils recovery.

As it was mentioned before, in the case of survival rule induction, there are no negative examples, therefore the precision of all rules was equal to 1. Similarly to the classification rules, the majority of survival rule sets generated by the guided induction were more stable (smaller standard deviation) than rule sets generated in automatic mode. Statistically, there were no differences between IBS values of guided and automatically generated sets of rules. The similarity measure values of rule sets were very low. This demonstrates that it is possible to define a rule set compliant with the user’s needs, which is different from the automatically generated model, but preserves its prediction abilities.

The next step in the doctor's analysis could be a deeper investigation of the induced rules. For example, our algorithm could be used for further analysis of the s-R1 rule. One can remove the conditions that may be too specific according to the medical knowledge (e.g. $ANCRecovery \leq15$) and analyze the quality of that modified rule---separately or together with another rules induced in an automatic way.

The presented example shows that the visualization of rule conclusions is very helpful in the survival analysis. Furthermore, similarly to the previous cases, an interactive analysis of data and induced rules rendered interesting results. The models showed better compliance with the user’s (e.g. doctor’s) requirements than those achieved by means of an automatic method.

\section{Conclusions}
\label{sec:conclusions}

The article presents a rule induction algorithm in which the learning process can be guided by the user (domain expert). GuideR can be used in classification, regression, and survival settings in an interactive way, enabling the user to adjust final rule set to his own preferences.
The rule induction algorithms are known to be unstable, as a small change in the set of the training examples may cause significant changes in the resulting rule set. The underlying cause is often ralated to the boundary areas of elementary conditions covering only small number of examples. A user-guided definition of those ranges usually results in preserving predictive abilities of the final rule set, making it more stable, clearer, and closer to the user’s intuition at the same time. For example, in the analysed case studies, the survival rule \textbf{\cmt}{s-R1} contained a condition with a range 13.055---limiting this range to 13 makes the rule more intuitive with insignificant decrease in the quality.

GuideR can impact the attributes, elementary conditions, or even rules of which the rule sets are composed of, directing the induction towards models most interesting to the user. Thus, the algorithm can be considered as a tool for knowledge discovery and for testing certain hypotheses concerning dependencies which are expected to occur in the data.
In particular, the algorithm is able to find modifications of user-defined hypotheses, provided in the form of rules, to improve their quality. 
Certainly, an automatic rule induction can be the starting point of a thorough dependency analysis. A set of automatically induced rules---or selected rules from this set---can be the basis for further, interactive experiments. Moreover, the guided induction can be an iterative process, i.e., the successive rule sets may be built on the basis of the insights from the previous iterations.

The efficiency of our algorithms for automatic rule induction has been confirmed on dozens of benchmark datasets \cite{wrobel2017,wrobel2016,sikora2012, sikora2013data}. In the experimental part of this article we focused on showing the efficiency and benefits coming from the use of the guided version of the algorithm. For this purpose, the analysis of three real-life datasets was presented. It show that the guided rule induction may produce data models of similar generalization abilities (e.g., classification accuracy) as the automatic induction, containing attributes, elementary conditions, and rules complying with the user’s requirements.


Further work will concern two directions. The first one is extending the algorithm with the possibility to induce so-called action rules [] and interventions. Action rules and interventions specify recommendations which should be taken in order to transfer objects from the undesirable concept to the desirable one (e.g., moving a client from the churn group to the group of regular customers). The second direction will be focused on the development of a graphical user interface for GuideR to make it easier to apply in the real-life analyses.

\section*{References}

\bibliography{references}

\end{document}